\documentclass[letterpaper, 10 pt, conference]{ieeeconf}  % Comment this line out if you need a4paper

\IEEEoverridecommandlockouts                              % This command is only needed if 
\usepackage{graphicx}
\usepackage{verbatim}
\usepackage{array}
\usepackage{upgreek}
\usepackage{float}   
\usepackage{amsmath,amssymb}
\usepackage{cite}
\usepackage{booktabs}
\usepackage{algorithm}
\usepackage[noend]{algpseudocode}

\usepackage{todonotes}
\usepackage{xcolor}

\usepackage[normalem]{ulem}
\let\labelindent\relax
\usepackage{enumitem}
\usepackage[hidelinks]{hyperref}
\makeatletter
\def\algbackskip{\hskip-\ALG@thistlm}
\makeatother

\usepackage{tikz}
\usetikzlibrary{positioning, shapes, arrows.meta}
\usetikzlibrary{shapes,arrows,chains}
\usetikzlibrary{arrows,calc,automata}

\tikzset{
    block/.style = {draw, rectangle, 
        minimum height=.5cm, 
        minimum width=1cm},
    input/.style = {coordinate,node distance=.5cm},
    output/.style = {coordinate,node distance=.5cm},
    arrow/.style={draw, -latex,node distance=.5cm},
    pinstyle/.style = {pin edge={latex-, black,node distance=1cm}},
    sum/.style = {draw, circle, node distance=1cm},
    line/.style={-{Stealth}}
    }

\title{\LARGE \bf
Design and Implementation of an Ultra-Low-Cost Wall-Climbing Robot for Infrastructure Crack Detection}

\author{
Mrinmoy Modak$^{1}$,
Supreyo Chakravorty Pretom$^{2}$,
Shourv Tarafder$^{2}$,
Daniel S. Drew$^{1}$%
\thanks{$^{1}$Department of Electrical and Computer Engineering,
University of Hawaii at Manoa, Honolulu, HI 96822, USA.}
\thanks{$^{2}$Department of Electrical and Electronic Engineering,
Chittagong University of Engineering \& Technology (CUET),
Chattogram 4349, Bangladesh.}
\thanks{Corresponding author: Daniel S. Drew,
\tt{ddrew@hawaii.edu}}
}

\begin{document}

\maketitle
\thispagestyle{empty}
\pagestyle{empty}

\begin{abstract}
   Crack detection is a crucial process to ensure the safety and longevity of buildings and other infrastructure. In this paper, we developed a low-cost, automated crack detection robot that leverages CNN, EfficientNet-B0, and YOLOv8 for efficient identification of cracks in concrete surfaces with a curated crack image dataset introduced to support training and evaluation. YOLOv8’s real-time object detection enhances crack localization, while CNN and EfficientNet-B0 provide binary classification, ensuring high precision and recall. The system consists of two stages. In the first stage, YOLOv8 detects and localizes wall regions from the video frame, and the bounding boxes are cropped. The second stage performs crack detection using one of three models by analyzing the cropped regions. Cost-effective approaches are also taken for robot design. The robot features a fan-based negative pressure adhesion system, a 4-wheeled skid-steering drive, and an ESP32-CAM for real-time image capture. Its lightweight 3D-printed chassis ensures stability, allowing it to navigate both walls and ceilings while capturing images for crack analysis. Unlike conventional wall-climbing robot designs, this robot incorporates a funnel-shaped body that enhances negative pressure generation and achieves a 44\% reduction in duty cycle, significantly lowering power consumption. By combining low-cost hardware with a deep learning pipeline, our system provides a scalable, efficient, and accessible solution for real-time infrastructure inspection at an approximate total cost of \$25, with a lightweight web application enabling smartphone-based control. This affordability makes the system more suitable for the developing world, where infrastructure inspection is often limited by budget constraints, labor intensity, and safety risks.
\end{abstract}

\section{Introduction}
\label{"Introduction"}
Civil infrastructure, including bridges, dams, and skyscrapers, becomes susceptible to structural failure as it deteriorates over time. This structural degradation raises serious safety concerns, which is why structural health monitoring has become a critical area of research \cite{50}. However, detecting cracks in vertical walls, ceilings, and narrow passages remains a challenging task due to accessibility and viewing constraints. Traditional inspections are subjective, labor-intensive, and potentially hazardous. Recent advancements in robotics and deep learning have enabled the development of autonomous systems for infrastructure monitoring, enhancing detection accuracy and inspection efficiency. 

\begin{figure}[t]
    \centering
    \includegraphics[width=01\linewidth]{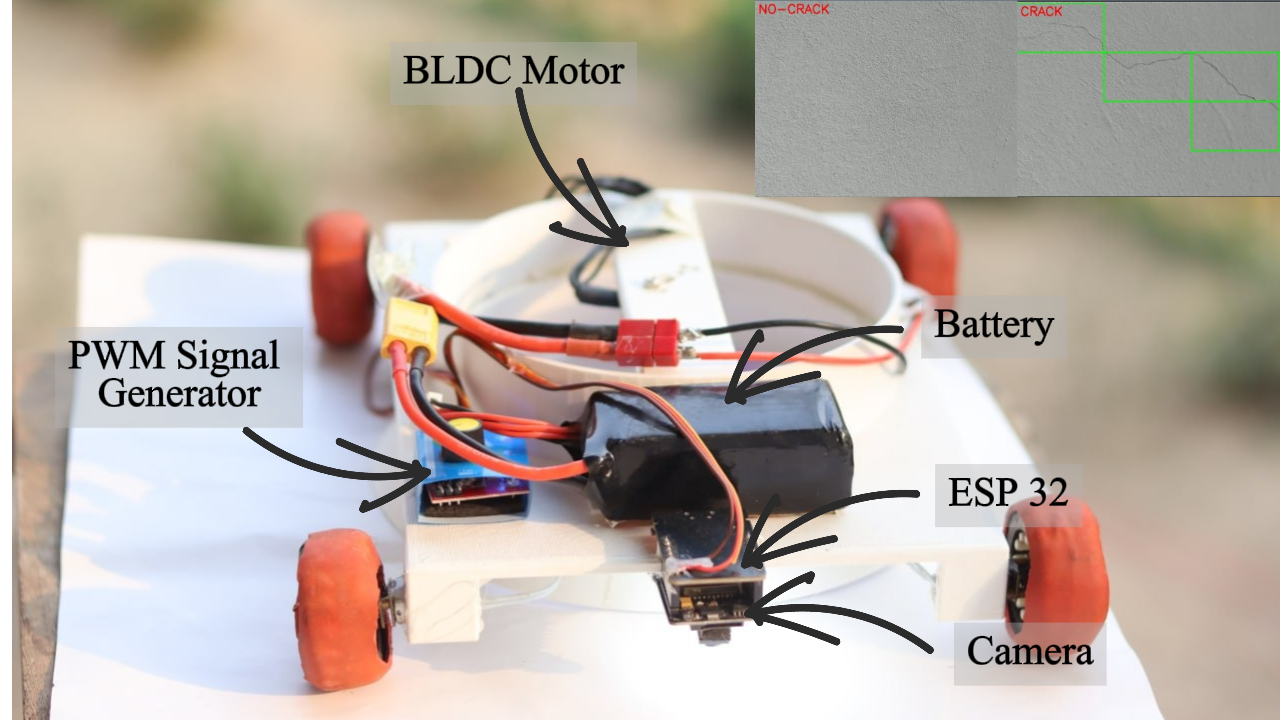}
    \caption{Ultra-low-cost wall-climbing robot for autonomous concrete crack detection using deep learning, with key hardware components labeled and a sample crack classification output (top-right).}
    \label{Poster Image}
\end{figure}

 A deep learning-based crack detection framework using a convolutional neural network (CNN) can be used to detect cracks in concrete surfaces \cite{19}. CNNs, along with other deep-learning methods, automatically learn discriminative crack features such as edges and crack textures while providing robustness against environment variations such as shadows, color, and surface non-homogeneity \cite{3}. Many of these systems, however, require high-end hardware and incur significant computational costs. Consequently, they fail to provide low-cost and efficient solutions for real-life applications. There is therefore a clear need for cost-effective robotic inspection systems that balance computational efficiency with reliable crack detection. 

The wall-climbing robot implemented in our paper, shown in Fig.~\ref{Poster Image}, employs a fan-based mechanism based on negative pressure. Negative pressure ensures secure adhesion to inclined surfaces, walls, and ceilings. A four-wheel drivetrain allows the robot to traverse vertical surfaces with high levels of maneuverability, while an ESP32-CAM provides real-time image and video capture for detailed inspection. Its lightweight 3D-printed chassis ensures stability and agility, even in complex environments. The video stream is then processed by deep-learning models, including YOLOv8, EfficientNet-B0, and a CNN, which detect cracks in the surface in real time.

This study presents three key contributions. First, we introduce a chassis that is expressly designed to enhance negative pressure effects in a passive manner. The funnel-shaped body geometry strengthens adhesion force and improves stability on inclined surfaces, vertical walls, and ceilings while preserving energy efficiency by reducing air leakage, increasing airflow velocity, and lowering internal static pressure. Second, we create an extremely inexpensive (approximately \$25) transportable robotic system that can record video in real time on these difficult terrains. In order to ensure practical deployment in resource-constrained situations, the system is constructed utilizing locally accessible, reasonably priced components, and incorporates lightweight deep-learning techniques suitable for consumer-grade computers. Third, we present an open-source collection of  40,172 labeled crack and non-crack photos taken under various lighting and material conditions, offering a publicly available standard for future infrastructure inspection studies.

\section{Background and Related Works}
\label{"Background and Related Works"}
\subsection{Wall-Climbing Robots}
Wall-climbing robots have attracted significant research interest due to their potential to perform inspection, maintenance, and monitoring tasks on vertical and inverted surfaces. Manual assessments and on-site inspections of cracks are challenging due to various limitations, including subjectivity, high labor costs, time consumption, and susceptibility to environmental conditions such as lighting and surface texture \cite{1}. These limitations necessitate the development of new robots that can detect cracks while operating reliably under varying environmental conditions, supporting automated real-time infrastructure monitoring \cite{2}.

A fundamental challenge in the implementation of wall-climbing robots lies in maintaining stable adhesion while simultaneously preserving controlled mobility. Adhesion performance and vehicle stability are greatly impacted by variations in surface characteristics, including texture, material composition, inclination angle, and friction coefficient. In order to guarantee dependable operation on a variety of surfaces, wall-climbing systems must be built to adapt to these shifting environmental conditions. Various adhesion strategies have been implemented, including ducted fan systems \cite{39}, self-compliant tracked mechanisms \cite{40}, and bio-inspired adhesion technologies \cite{31}. These adhesion techniques enable each of these robots to navigate on wall surfaces with various textures.

Wall-climbing robots require not only dependable adhesion mechanisms but also robust sensing and control systems to preserve functionality across a variety of surface conditions. Environmental disturbances such as vibration, optical distortion, and lighting variation, for example, can severely degrade inspection accuracy \cite{9}. To address these critical challenges, control architectures that incorporate stability, localization, and mapping are increasingly emphasized to ensure reliable and coordinated wall-climbing operations. The study presented in \cite{10} demonstrated real-time navigation with camera pose alignment, allowing robots to accurately localize structural defects. Key techniques, such as behavior-based control strategies \cite{33} and path-finding technologies \cite{25}, are used to enhance operational safety. Moreover, drone-assisted inspection approaches have been deployed to evaluate wall surfaces \cite{36}, alongside rock-climbing robotic platforms built for vertical exploration and assessment \cite{42}.

\subsection{Deep Learning Approaches for Crack Detection}

Deep learning techniques detect cracks on wall surfaces, particularly in challenging environments with variable textures, colors, and lighting conditions. Among these techniques, convolutional neural networks (CNNs) are widely applied. This is largely because hybrid dilated convolution architectures have demonstrated strong pixel-level segmentation performance in complex backdrops by effectively capturing multi-scale contextual information \cite{6}. UAV-assisted CNN frameworks further enhance detection capability by enabling efficient data collection over large areas \cite{7}. Recent architectures, such as ConvNeXt V2-based models, improve edge delineation and feature representation compared to older CNN techniques, resulting in more precise fracture boundary identification on fair-faced concrete surfaces \cite{18}. Attention-enhanced CNN frameworks also increase sensitivity, enabling the reliable detection of tiny cracks that would otherwise be difficult to identify \cite{17}. However, the CNN has a major downside: it needs a large amount of data to predict accurately. Data scarcity can become a major problem when applying a CNN. For this reason, techniques such as synthetic data augmentation \cite{19} and super-resolution preprocessing \cite{20} mitigate these issues but can increase computational demands. Additional strategies include hybrid supervised–unsupervised learning for fine-crack detection \cite{23}, specialized super-resolution networks for micro-crack enhancement \cite{24}, CNN-based segmentation models \cite{27}, and machine-learning-based crack-classification frameworks \cite{26}. These approaches improve detection robustness in challenging conditions such as shadows and low light. Nevertheless, many of these techniques depend on high-quality input data or computationally demanding models, which may restrict their usefulness in real-time or resource-constrained situations.

Concurrently, the real-time performance of YOLO-based object identification systems has drawn attention \cite{10}. Fast and precise fracture identification that is appropriate for field inspections is made possible by improved YOLO architectures. While depth-assisted YOLO systems increase robustness in industrial settings \cite{14}, integrating RGB-D imaging with ground-penetrating radar (GPR) in YOLO pipelines enables simultaneous surface and subsurface fault investigation \cite{9}. 
 
Existing wall-climbing robotic systems have shown promising capabilities for inspection and navigation on vertical and sloped surfaces, as well as a range of adhesion techniques, such as magnetic and bio-inspired approaches. Several platforms integrate vision systems and autonomous control strategies for surface assessment tasks. However, many of these systems rely on expensive sensors and processing units, high-power actuation, or intricate mechanical designs, which restrict scalability and cost-effective deployment, especially in large-scale infrastructure inspection scenarios. Furthermore, comparatively few studies combine energy-conscious design, economically scalable sensory systems, and passive adhesion efficiency into a single framework. The present work builds on a negative-pressure approach within a deliberately low-cost and deployable inspection architecture, thereby complementing existing solutions while addressing practical accessibility and scalability constraints.
\section{System Architecture and Mechanism}

\begin{figure}
    \centering
    \includegraphics[width=\linewidth]{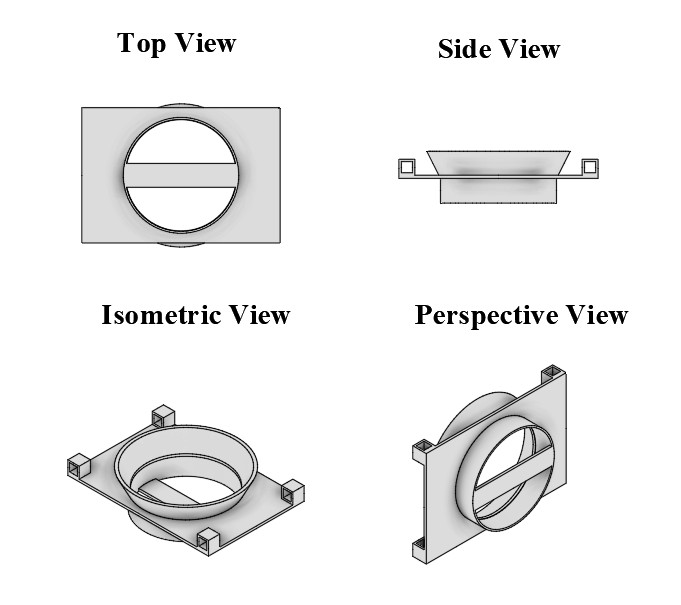}
    \caption{CAD model of the wall-climbing robot chassis showing the top, side, isometric, and perspective views of the funnel-shaped structure.}
    \label{fig:cad_model_reference_frame}
\end{figure}

The wall crack detection system utilizes a specialized robot equipped with a fan-based thrust adhesion system, a 4-wheeled drive, an ESP32-CAM for real-time image and video capture, an L298N motor driver for motor control, and a lightweight chassis frame. The remote control driving system enables controlled movement, while the ESP32-CAM captures high-resolution images for crack analysis. The lightweight chassis ensures stability during operation, making the robot suitable for crack detection in various environments. Fig.~\ref{fig:cad_model_reference_frame}  shows the robot chassis from top, side, isometric, and perspective views.

To adhere to the wall, the robot uses a high-speed propeller fan to generate negative pressure. As shown in Fig.~\ref{fig: Negative Pressure absorption mechanism}, ambient air enters the chassis through side inlets and is directed toward a centrally mounted fan.  Air is then expelled by a high-speed propeller fan upward, and this reduces the internal pressure beneath the robot relative to the surrounding atmospheric pressure. This differential pressure generates a normal force that pushes the robot against the wall surface. This mechanism is enhanced by a funnel-shaped printed inlet. The ambient air that enters through the lower side of the robot is guided through the convergent funnel toward the impeller. The narrowing geometry accelerates the airflow, reducing the internal pressure further, and this also decreases air leakage. The robot utilizes rubber tires to enhance traction on vertical surfaces. The robot's lightweight 3D-printed chassis, fabricated from PLA, minimizes weight while ensuring thrust support. Central clearance is provided for airflow and camera coverage, with the center of mass positioned near the fan to enhance stability. The complete mechanical design files are available at:  \url{https://github.com/thedrl/wall-climbing-robot/tree/main/3D_Models} 

\begin{figure}[t]
    \centering
    \includegraphics[width=1\linewidth]{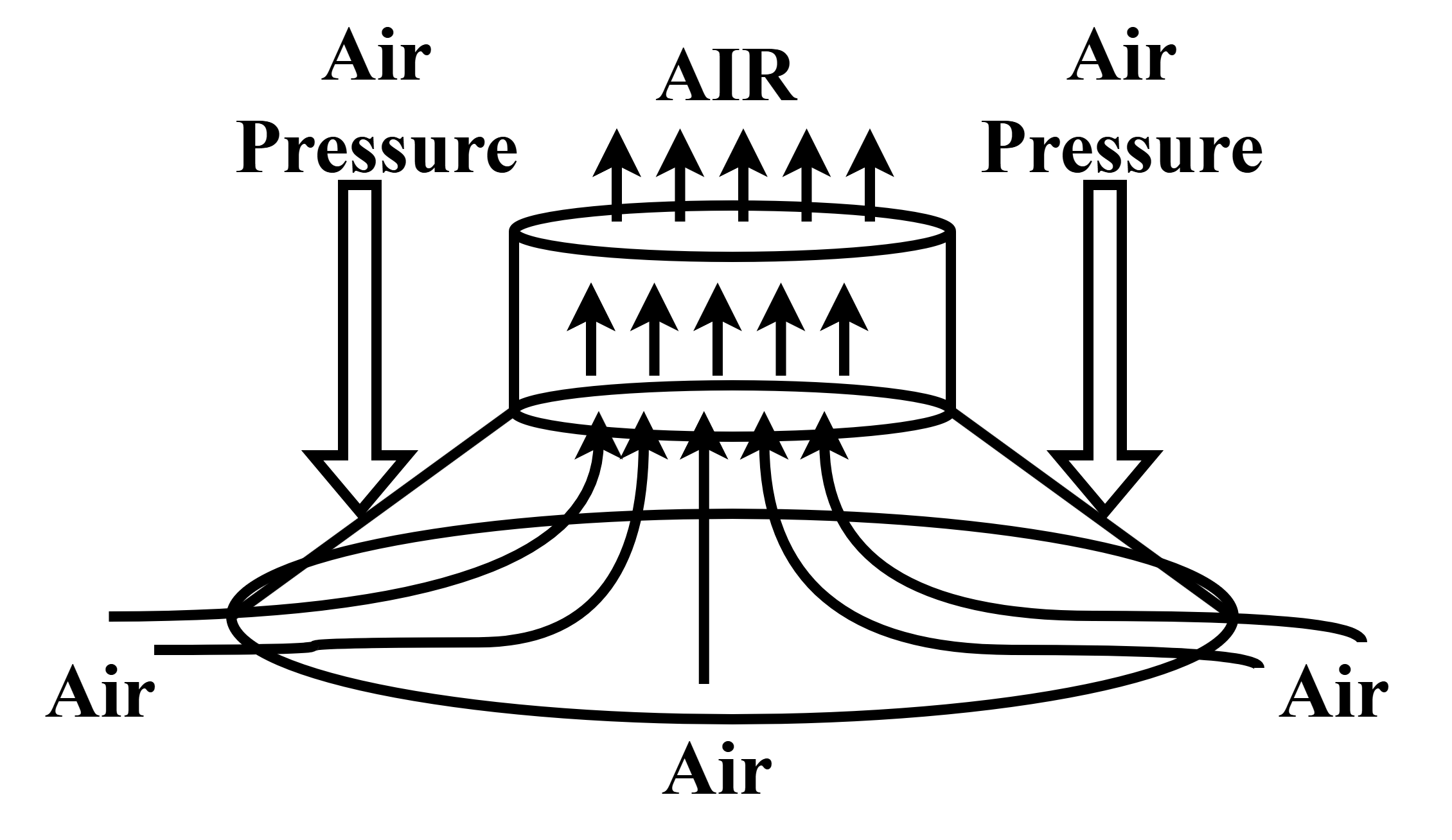}
    \caption{Illustration of the negative pressure adhesion mechanism, where upward airflow creates a low-pressure region inside the chamber that presses the chassis against the surface.}
    \label{fig: Negative Pressure absorption mechanism}
\end{figure}

\section{Deep Learning System Architecture}
A deep learning-based pipeline is used to improve the robot's capacity to automatically identify cracks on concrete surfaces. Images were gathered,
enhanced, and preprocessed in order to train this pipeline. First, a single-class YOLOv8 model is trained to identify wall surfaces using the images. After that, the dataset is shuffled and separated into categories for wall surfaces with cracks and those without. Then, to differentiate between crack and no-crack images, binary detection models (YOLOv8, CNN, and EfficientNet-B0) are used. In order to determine the best architecture for accurate crack detection, each model's performance is assessed.

 \subsection{Structural Wall Crack Detection Dataset}
   
\begin{figure}[h]
    \centering
    \includegraphics[width=1\linewidth]{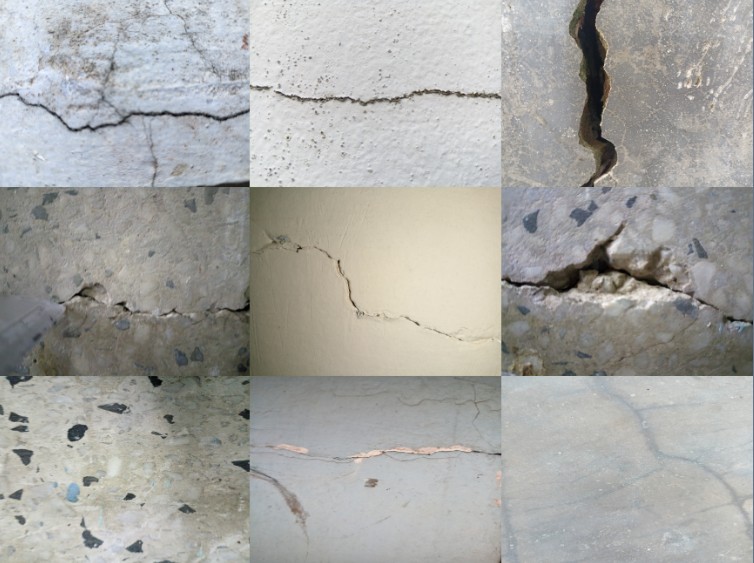}
    \caption{Sample images from the surface crack dataset showing variations in crack width, orientation, texture, and lighting conditions.}
    \label{fig: Dataset for Surface Crack Detection}
\end{figure}

To identify walls with cracks and without cracks, images were collected representing both cracked and crack-free wall surfaces. Initially, the dataset was sourced from Kaggle; however, it lacked diversity in terms of wall color and condition, making it insufficient for training a deep-learning model with high accuracy. To address this limitation, additional images, as shown in Fig.~\ref{fig: Dataset for Surface Crack Detection}, were collected from the campus of Chittagong University of Engineering and Technology (CUET), Bangladesh.

\subsection{Data Augmentation and Preprocessing}
The images were collected from both online sources and direct data collection efforts. The dataset was divided into training, test, and validation sets using a 70:15:15 ratio. To enhance diversity and make the dataset more applicable to real-world scenarios,  several augmentation techniques were applied to the images, including scaling, rotating, flipping, adjusting contrast, RGB-based data augmentation techniques, and cropping. These resulted in a total of 40,172 images, where 20,050 images are of cracked wall surfaces and 20,122 images are of crack-free wall surfaces. The images were then resized to 180 by 180 pixels. Here, augmentation was done after the dataset split to prevent data leakage. The dataset is made openly available at \url {https://github.com/thedrl/wall-climbing-robot} to support reproducibility and further research in low-cost infrastructure inspection.

\subsection{YOLOv8 (You Only Look Once, Version 8)}
YOLOv8, a recent iteration of Ultralytics' You Only Look Once (YOLO) family, is one of the major developments in one-stage object detection frameworks. This technique uses an anchor-free detection head and an updated backbone to directly predict object classes and bounding boxes without the need for predefined anchors. This design facilitates training, increases generalization across various datasets, and improves localization accuracy. YOLOv8 is well-suited for real-time applications because it is optimized for high-speed inference while retaining strong detection performance \cite{41}. In comparison to previous YOLO variants and other state-of-the-art detectors, recent studies have shown its efficacy across a wide range of computer vision tasks, achieving a balance between detection accuracy and inference latency even with limited training data \cite{30}.

\subsubsection{Single Object Detection through YOLOv8}
YOLOv8 is used, specifically its single object detection, to identify walls in front of the camera. Images of walls, including both crack and non-crack surfaces, are used to implement the YOLOv8 single object detection. For the image's matching annotation labels, another directory is made. Both the label files and the image dataset are split into subsets for training and validation. The YOLO format is used to prepare annotations, and each label file includes the class ID that represents the wall object along with the corresponding bounding box details, such as the normalized x-center, y-center, width, and height. Dataset configuration and class definitions are specified using YAML files. All images are resized to 644 × 644 pixels, which ensures uniform input dimensions, and the model is trained for 10 epochs. This single-object detection model enables the YOLOv8 to detect the wall surface efficiently.

\subsubsection{Binary Detection through YOLOv8}
For crack detection using YOLOv8, the images, previously divided into training, test, and validation sets, are categorized as cracked or non-cracked wall surfaces and resized to 644 × 644 pixels to ensure a consistent input size for YOLOv8. YOLOv8 is applied for object classification, and the model is trained for 10 epochs. The network classifies the images into cracked or non-cracked categories for crack detection.

\subsection{Convolutional Neural Networks (CNN)}
 CNNs are based on a mathematical model of convolution. Convolution employs kernels or filters on the feature map and produces output feature maps.
The CNN architecture has two stages: one for extracting features and the other for classification. The input image with size 180×180×3 pixels is processed through a series of convolutional layers with Rectified Linear Unit (ReLU) activation functions. These layers learn different spatial features/dimensions and progressively reduce the spatial dimensions, producing a final feature-map size of 45 × 45 × 64. The network can automatically learn hierarchical representations of visual patterns thanks to the feature extraction stage. While deeper layers encode more abstract and high-level representations, convolutional filters move across the input image to capture low-level features, such as edges, textures, and shapes. ReLU activation increases learning efficiency and speeds up convergence by introducing non-linearity.

L2 regularization is used on the convolutional layers to improve generalization and reduce overfitting. 
\begin{figure}[t]
    \centering
        \centering
        \includegraphics[width=\linewidth]{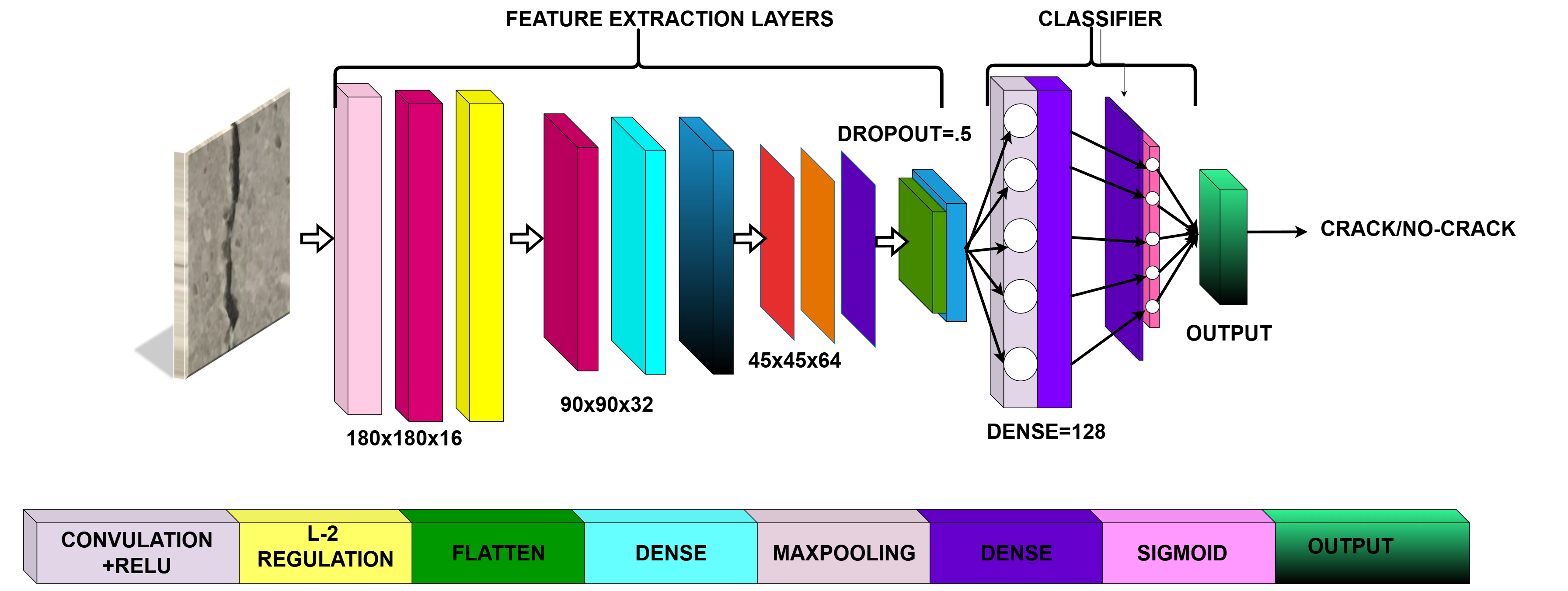}
        \caption{CNN architecture for crack detection, showing convolutional, pooling, and dense layers followed by a sigmoid output for binary classification.}
        \label{CNNarchi.jpg} 
\end{figure}
The three-dimensional feature maps (height, width, and depth) are transformed into a one-dimensional feature vector by a Flatten layer after feature extraction. The classification stage, which uses a fully connected Dense layer with 128 neurons to interpret the learned high-level features, then receives this vector. A sigmoid activation function is used in the final output layer to generate class probabilities for both cracked and non-cracked images. The network is trained for 25 epochs, which was found to provide stable convergence and reliable classification performance. Fig.~\ref{CNNarchi.jpg} shows the architecture of the CNN model.

\subsection{ EfficientNet-B0}EfficientNet utilizes a coefficient to uniformly alter the depth, width, and resolution of the models. Notably, the baseline EfficientNet-B0 variant strikes a balanced trade-off between model complexity and performance, making it well-suited for a diverse array of computer vision tasks, particularly those with restricted computational resources \cite{37}. In our methodology, EfficientNet-B0 is employed as the backbone feature extractor for binary crack classification due to its ability to achieve high accuracies with limited training data. The pre-trained EfficientNet-B0 model, initialized with ImageNet weights, is used without its top classification layers. Input images are resized to 224×224 pixels. Only the recently added classification layers can be trained during the first training phase since all EfficientNet-B0 layers are frozen to maintain the learned generic visual features. A Global Average Pooling layer is used to process the extracted deep feature maps, reducing spatial dimensions while maintaining discriminative information. To reduce overfitting, two dropout layers with a rate of 0.3 and 0.4 are used. A single neuron with a sigmoid activation function makes up the final classification layer, which allows for binary classification between cracked and non-cracked images. After prior convergence, a fine-tuning strategy is adopted by unfreezing the last 20 layers of the backbone. This allows the higher-level feature representations to adapt more closely to the crack detection task while keeping the majority of the network frozen to prevent overfitting. Fine-tuning is performed using a lower learning rate to ensure stable weight updates.

\subsection{Video Processing}

Each frame from the camera is analyzed independently. Each frame is first divided into a 2×2 grid of non-overlapping patches, enabling localized analysis of different regions within the frame. The patches are resized to 224 × 224 pixels for EfficientNet-B0, 180 × 180 pixels for CNN, and 644 × 644 pixels for YOLOv8. First, these frames are analyzed by single object detection of YOLOv8, then if no wall is present in the frame on the screen, ‘No Wall’ is shown. However, if there is a wall, this frame is then analyzed by the full YOLOv8, CNN, and EfficientNet pipeline. The recovered frames are preprocessed using digital zoom, color space conversion from BGR to RGB, and pixel value normalization in order to enhance the performance of the binary classification models.

A patch is classified as cracked if its predicted probability falls below a threshold of 0.5, and the corresponding patch coordinates are recorded for localization. If one of the four patches is classified as a crack, the video shows it as a ‘Crack’ with a labeled bounding box highlighting the cracked region. 

To increase the speed in real-time use, the YOLOv8 wall detector analyzes once every 3rd frame; however, the classifier models analyze every frame that is passed by YOLOv8 as a wall.

\section{Result Analysis}
\subsection{Crack Detection By YOLOv8 Results}

\begin{figure}[htbp]
    \centering
    
        \centering
        \includegraphics[width=\linewidth]{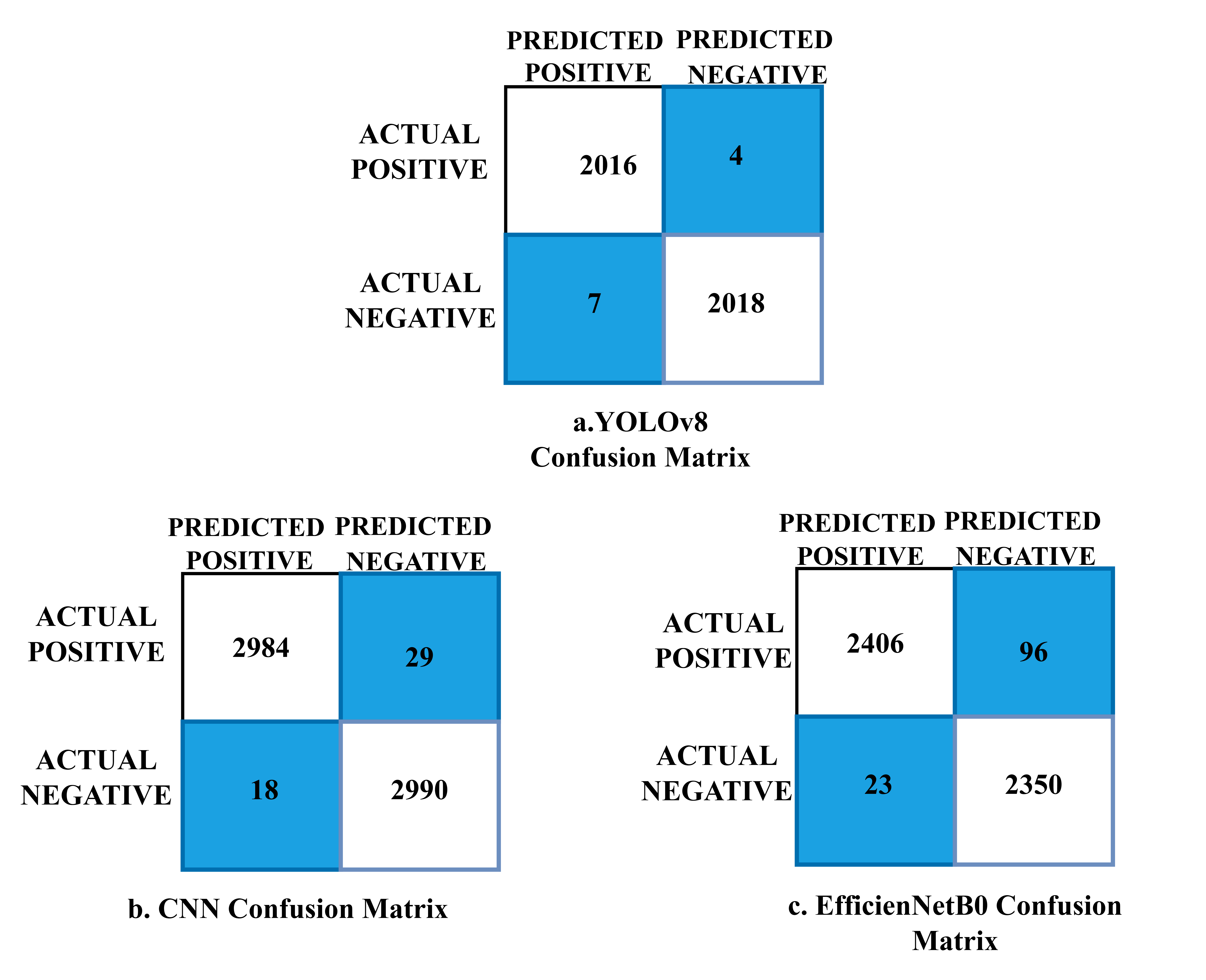}
        \caption{a. Confusion matrix of YOLOv8, b. Confusion matrix of CNN, c. Confusion matrix of EfficientNet-B0.}
        \label{Confusion Matrix}
\end{figure}

Confusion Matrix for Crack Detection: The confusion matrix, as shown in Fig.~\ref{Confusion Matrix}(a), illustrates the performance of the YOLOv8 model, which achieved 2016 true positives and only 4 false negatives, 2018 true negatives, and merely 7 false positives, which demonstrates highly reliable crack detection performance.

\begin{figure*}[h]
    \centering
 
        \centering
        \includegraphics[width=0.9\linewidth]{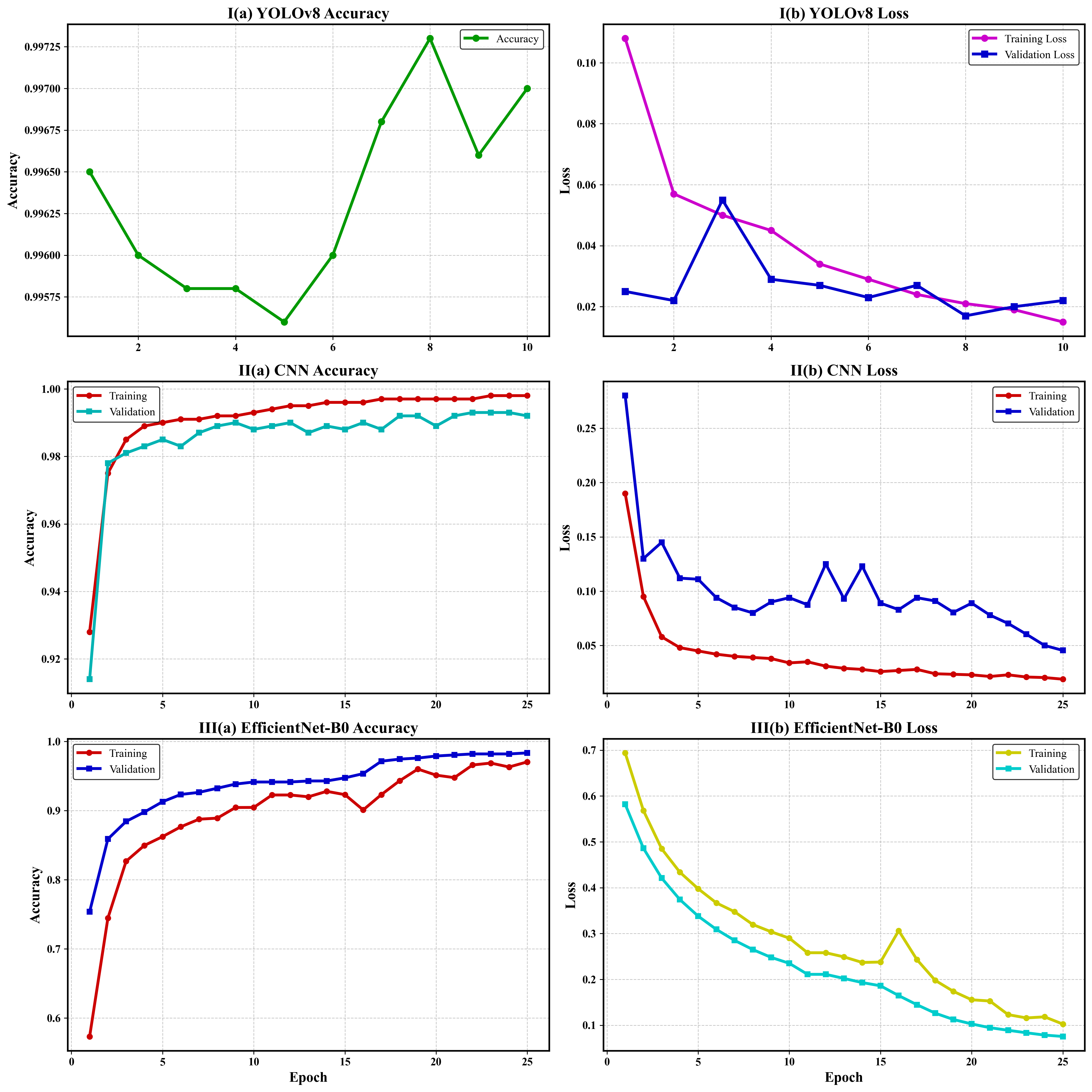}
        \caption{Training performance comparison of the three deep learning models. (I(a)) YOLOv8 accuracy, (I(b)) YOLOv8 loss, (II(a)) CNN accuracy, (II(b)) CNN loss, (III(a)) EfficientNet-B0 accuracy, and (III(b)) EfficientNet-B0 loss over the training epochs.}
        \label{comparison_yolo_cnn_efficientnet_styled (1)}
        \vspace{-1em}
\end{figure*}

 Training and Validation Loss: The graph, as shown in Fig.~\ref{comparison_yolo_cnn_efficientnet_styled (1)}, shows the loss and accuracy curve for training and validation. A rapid drop in training loss indicates rapid learning, while validation loss fluctuations suggest slight overfitting, as shown in Fig.~\ref{comparison_yolo_cnn_efficientnet_styled (1)}(I)(b). Fig.~\ref{comparison_yolo_cnn_efficientnet_styled (1)}(I)(a) shows the accuracy over the epochs.

 Table~\ref{tab:recall_precision_f1_comparison}(a) shows the recall, precision, and F1-score of the YOLOv8.
\begin{table}[h!]
\centering
\caption{Recall, precision, and F1-score comparison of crack detection models.}
\label{tab:recall_precision_f1_comparison}
\begin{tabular}{|c|c|c|c|}
\hline
\textbf{Model} & \textbf{Recall (\%)} & \textbf{Precision (\%)} & \textbf{F1-score (\%)} \\ \hline
\textbf{(a) YOLOv8} & 99.80 & 99.65 & 99.72 \\ \hline
\textbf{(b) CNN} & 99.04 & 99.40 & 99.22 \\ \hline
\textbf{(c) EfficientNet-B0} & 96.16 & 99.05 & 97.59 \\ \hline
\end{tabular}
\end{table}

\subsection{Crack detection by CNN Results}
Confusion matrix: The confusion matrix, as shown in Fig.~\ref{Confusion Matrix}(b), displays the CNN model's performance in crack detection. The model correctly identifies most cracks (2984 true positives) and non-cracked areas (2990 true negatives), with few misclassifications (29 false negatives, 18 false positives).

Accuracy vs. Epochs: Fig.~\ref{comparison_yolo_cnn_efficientnet_styled (1)}(II)(a)  shows the accuracy of training and validation, while Fig.~\ref{comparison_yolo_cnn_efficientnet_styled (1)}(II)(b) shows the loss (training loss in red, validation loss in blue). Here, a fluctuation of validation loss can also be observed, though the training loss declines smoothly. This also suggests there is a slight overfitting present in the model, though the accuracy remains high. L2 regularization compensates for some overfitting, but the lower quality and variations of the data prevent the validation loss curve from fully stabilizing. 
Table~\ref{tab:recall_precision_f1_comparison}(b) shows the recall, precision, and F1-score of the CNN model.

\subsection{Crack detection by EfficientNet-B0 Results}
Confusion matrix: The confusion matrix, as shown in Fig.~\ref{Confusion Matrix}(c), displays the  model's performance in crack detection. The model correctly identifies most cracks (2406 true positives) and non-cracked areas (2350 true negatives), with few misclassifications (96 false negatives, 23 false positives).

Accuracy vs. Epochs: Fig.~\ref{comparison_yolo_cnn_efficientnet_styled (1)}(III)(a)  shows the accuracy of training and validation, while Fig.~\ref{comparison_yolo_cnn_efficientnet_styled (1)}(III)(b) shows the loss (training loss in red, validation loss in blue). Here, smooth declines in both validation and training losses are observed. Moreover, the accuracy remains high. This model shows less overfitting, although the dataset has limited variation.   
Table~\ref{tab:recall_precision_f1_comparison}(c) shows the recall, precision, and F1-score of the EfficientNet-B0 model. The complete implementation codes are available at: \url {https://github.com/thedrl/wall-climbing-robot}

\begin{figure} 
    \centering
    
        \centering
        \includegraphics[width=1\linewidth]{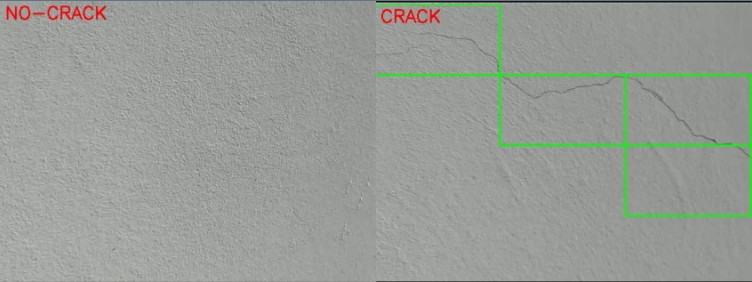}
        \caption{Real-time crack detection on wall surfaces. The proposed system correctly classifies non-crack and crack surfaces and localizes crack regions using green bounding boxes.}
        \label{/Wallnocrackwhite} 
\end{figure}

\begin{figure*}
      \centering
        \includegraphics[width=1\linewidth]{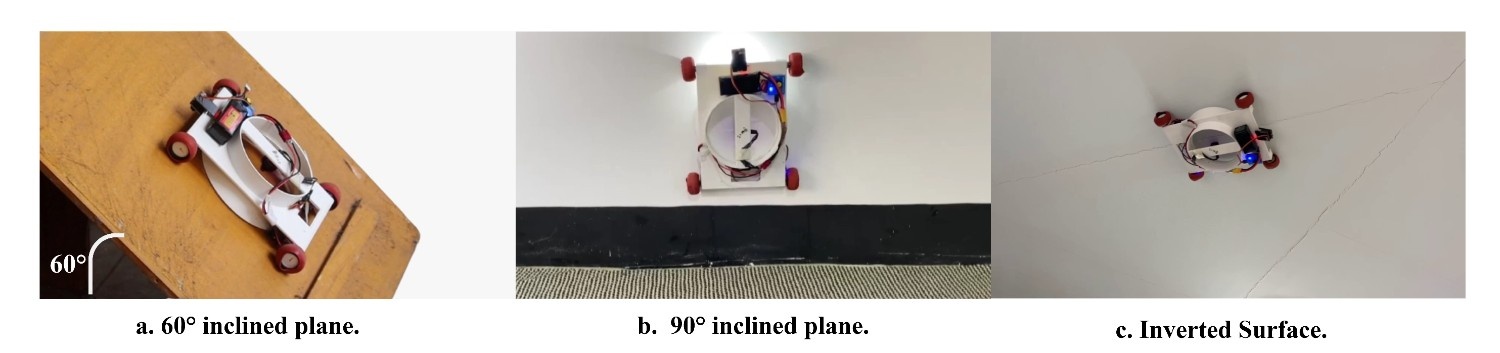}
        \caption{Experimental demonstration of the wall-climbing robot operating on different surface orientations: (a) stable locomotion on a 60° inclined plane, (b) vertical (90°) surface climbing, demonstrating sufficient negative pressure generation and stability against full body weight; and (c) inverted (ceiling) operation, confirming sustained adhesion and controlled motion while completely opposing gravity.}
        \label{/robosurface} 
\end{figure*}

\subsection{Real-time Crack Analysis}
The ESP32-CAM captures real-time video, and this video stream is transmitted to the computer through the Wi-Fi network connection. The computer then processes each frame of the received video using the crack detection pipeline. If a crack in the wall surface is detected, bounding boxes are drawn around the detected cracks, and the screen displays ‘Crack’. Otherwise, the screen displays ‘No Crack.’ Fig.~\ref{/Wallnocrackwhite} shows real-time crack detection.

\section{Mobility Assessment on Inclined and Inverted Surfaces}

The operator controls the robot through a mobile device. The robot demonstrates mobility on surfaces at various orientations, as shown in Fig.~\ref{/robosurface}. A web-based IoT control framework allows for remote monitoring and operation through a browser interface, ensuring platform-independent access to the robotic system. Sensor data is sent to either a cloud or a local server, where control commands are analyzed and communicated in real-time. This method improves system flexibility, scalability, and user access, while facilitating accurate and quick motion control. For inverted surfaces, compared to the same chassis without the funnel, the funnel-shaped shroud reduces the duty cycle from 90\% to about 50\%, corresponding to an approximate 44\% reduction in duty cycle and a significant decrease in power consumption.

\section{Cost Analysis}

\begin{table}
\caption{Hardware Components and Cost Breakdown.}
\label{tab:cost_breakdown}
\centering
\small
\setlength{\tabcolsep}{4pt}
\begin{tabular}{|l|c|r|}
\hline
\textbf{Component} & \textbf{Qty.} & \textbf{Cost (USD)} \\ \hline
BLDC Motor (RS2205) & 1 & 5.59 \\ \hline
Motor Driver (L298N) & 1 & 2.06 \\ \hline
Servo Controller & 1 & 1.65 \\ \hline
ESC (30A) & 1 & 4.12 \\ \hline
NodeMCU ESP8266 & 1 & 3.70 \\ \hline
ESP32-CAM & 1 & 3.70 \\ \hline
LiPo Battery (3.7V) & 3 & 4.32 \\ \hline
\multicolumn{2}{|r|}{\textbf{Total Cost}} & \textbf{25.14} \\ \hline

\end{tabular}
\end{table}

The cost breakdown for building the robot is shown in Table~\ref{tab:cost_breakdown}. Additionally, Table~\ref{tab:cost_comparison} shows the economic benefit of the suggested design by comparing the overall cost of the suggested robot to comparable robots documented in prior research. Based on the cost analysis, it can be concluded that the robot is highly cost-efficient and suitable for real-life applications in the developing world. 

The cost-effectiveness of the robot can be attributed to several design choices. Firstly, the usage of commercially available products in local markets. Secondly, using the adhesion mechanism of negative pressure with a funnel-shaped body structure that minimizes the need for high-power vacuum pumps or complex sealing mechanisms, thereby further reducing overall system cost. The bill of materials (BOM) is available at:  \url{https://github.com/thedrl/wall-climbing-robot/tree/main/BOM}

\begin{table}
\caption{Cost Comparison with Existing Wall-Climbing Robots.}
\label{tab:cost_comparison}
\centering
\small
\setlength{\tabcolsep}{4pt}
\begin{tabular}{|c|p{4.2cm}|r|}
\hline
\textbf{Ref.} & \textbf{Project} & \textbf{Cost (USD)} \\ \hline
\cite{47} & Wallybot wall-climbing robot & 825.00 \\ \hline
\cite{46} & Design and development of a wall-climbing robot for wall
inspection and crack detection & 175.32 \\ \hline
\cite{48} & Electromagnetic wall-climbing robot & 91.00 \\ \hline
\textbf{This Work} & \textbf{Vision-based crack detection robot} & \textbf{25.14} \\ \hline

\end{tabular}
\end{table}

\section{Future Work }
The suggested pipeline can be improved in the future by incorporating sophisticated deep learning architectures like ResNet50 for better feature extraction and U-Net for accurate fracture-region segmentation. To enhance detection performance, especially in difficult situations, these should be investigated as additions to or replacements for the current models in the pipeline.

Additionally, the system can be integrated with robotic automation, where inspection robots with the described crack-detection pipeline could use RRT* or other path-planning algorithms for autonomous traversal. By enabling real-time crack detection during autonomous navigation, this integration would completely automate the inspection procedure and qualify the system for extensive infrastructure monitoring.

\section{Conclusion}
  The findings demonstrate the enormous potential of incorporating deep learning into automated infrastructure inspection systems, where the funnel-shaped geometry reduces duty cycle by about 44\% while increasing airflow velocity and minimizing air leakage to enable steady vertical adhesion. A two-stage detection pipeline is suggested to further improve efficiency and lower deployment costs. The first stage involves locating and identifying wall regions within image or video streams using a YOLOv8 single-object detection model. In the second stage, the detected wall surface regions are further analyzed for cracks using detection models. EfficientNet-B0 outperforms both YOLOv8 and traditional CNN models in terms of overall performance. More accurate feature learning and localized analysis are made possible by segmenting images into patches. This modular approach significantly lowers computational complexity by limiting the detection process to pertinent regions only, making the system more effective. Moreover, the use of readily available, low-cost components without compromising performance makes our system cost-efficient. The framework offers a cost-effective and practical solution for infrastructure health monitoring, particularly in countries with developing economies where most of the other options are relatively expensive and unaffordable.
  
\bibliographystyle{IEEEtran}

\bibliography{reference}

\end{document}